\documentclass{INTERSPEECH2023}

\usepackage{amsmath,graphicx}
\usepackage{url}

\interspeechcameraready

\title{Leveraging Pretrained ASR Encoders for Effective and Efficient End-to-End Speech Intent Classification and Slot Filling}

\name{He Huang, Jagadeesh Balam, Boris Ginsburg}
\email{\{heh,jbalam,bginsburg\}@nvidia.com} 

\address{NVIDIA, USA}
\begin{document}
%
\maketitle
\begin{abstract}

We study speech intent classification and slot filling (SICSF) by proposing to use an encoder pretrained on speech recognition (ASR) to initialize an end-to-end (E2E) Conformer-Transformer model, which achieves the new state-of-the-art results on the SLURP dataset, with 90.14\% intent accuracy and 82.27\% SLURP-F1. We compare our model with encoders pretrained on self-supervised learning (SSL), and show that ASR pretraining is much more effective than SSL for SICSF. To explore parameter efficiency, we freeze the encoder and add Adapter modules, and show that parameter efficiency is only achievable with an ASR-pretrained encoder, while the SSL encoder needs full finetuning to achieve comparable results. In addition, we provide an in-depth comparison on end-to-end models versus cascading models (ASR+NLU), and show that E2E models are better than cascaded models unless an oracle ASR model is provided. Last but not least, our model is the first E2E model that achieves the same performance as cascading models with oracle ASR. Code, checkpoints and configs are  available.\footnote{\url{https://github.com/NVIDIA/NeMo/tree/main/examples/slu/speech_intent_slot}}

\end{abstract}
%
%

\section{Introduction}

Spoken language understanding (SLU) is an essential component of conversational AI, which aims to extract  semantic information directly from speech data. SLU has a very broad scope, including  intent classification~\cite{sharma2021intent}, slot filling~\cite{slurp,wang2021fine,arora2022espnet,seo2022integration}, speech emotion recognition~\cite{chen2020large,shon2021leveraging}, question answering~\cite{lee2018odsqa, you2021knowledge}, etc. There are mainly two types of SLU models: (1) cascading (ASR+NLU) models that first perform automatic speech recognition (ASR) and then apply a natural language understanding (NLU) model to the transcribed text; (2) end-to-end (E2E) models that directly predict the semantic output without predicting transcriptions. Compared with its counterpart natural language understanding (NLU), SLU is more challenging. On one hand, errors will propagate from ASR to NLU in cascading SLU models. On the other hand, end-to-end SLU models cannot make use of the pretrained large language models such as BERT~\cite{devlin2018bert}. To tackle these challenges, we use end-to-end SLU models that do not have error propagation, and we also show that better performance can be achieved by utilizing ASR encoders pretrained on out-of-domain datasets.

\begin{figure}[t]
    \centering
    \includegraphics[width=1.0\linewidth]{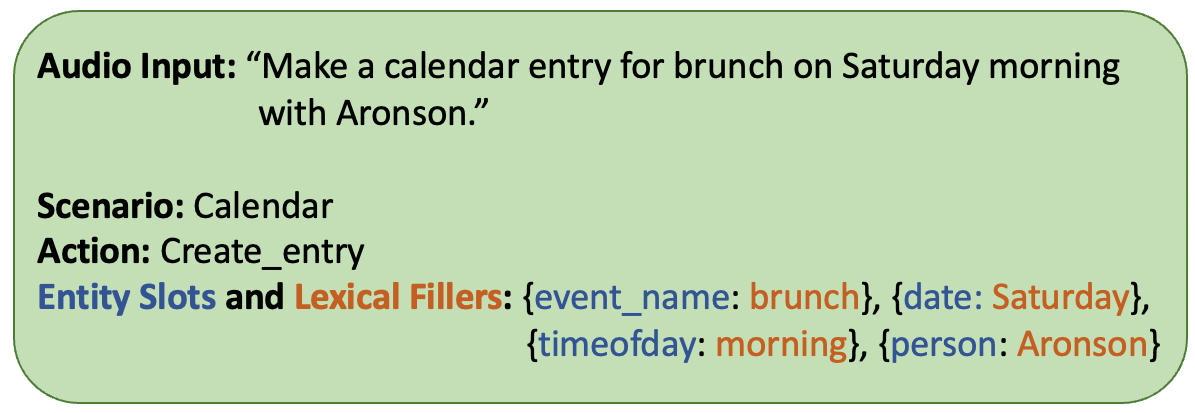}
    \caption{An example of speech intent classification and slot filling. An intent is composed of a \emph{scenario} and an \emph{action}, while \emph{slots} and \emph{fillers} are represented by \emph{key-value} pairs.}
    \label{fig:example}
\end{figure}

In this paper, we study the \emph{speech intent classification and slot filling} (SICSF) task, which aims to detect user intents and extract the corresponding lexical fillers for detected entity slots at the same time, as illustrated in Figure~\ref{fig:example}. The most common end-to-end approach in this task is the encoder-decoder framework, where the encoder is responsible for extracting acoustic features from input audios, and the decoder is responsible for decoding the features to semantic output of intents and slots. Current works~\cite{wang2021fine,seo2022integration} use encoders pretrained by self-supervised learning (SSL)~\cite{baevski2020wav2vec,hsu2021hubert}. However, there is a large domain gap between the self-supervised learning task and the SICSF task, which limits the benefits that SLU models can obtain from the SSL-pretrained encoders. Some recent works~\cite{arora2022espnet,seo2022integration} also propose to train the SLU model with additional ASR task in a multi-task loss, which is more tricky to train since it's hard to choose a proper weight to balance different loss terms. Also, such multi-task approach wastes some network parameters in learning the ASR decoder which is not used during inference phase of the SLU task.

\begin{figure*}[!ht]
    \centering
    \includegraphics[width=0.9\linewidth]{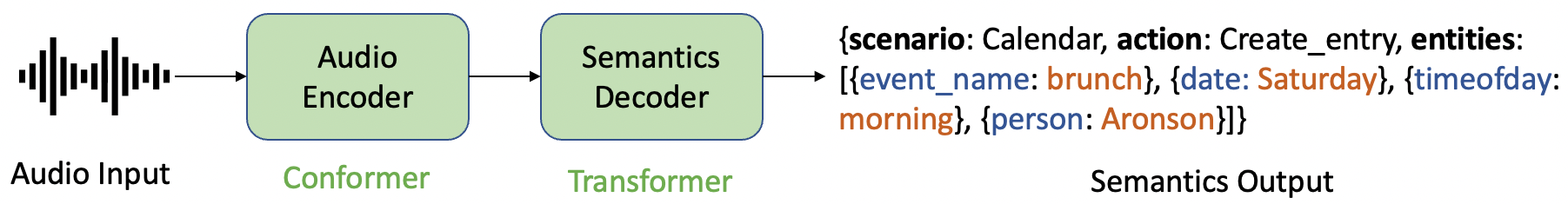}
    \caption{The proposed model for \emph{speech intent classification and slot filling} (SICSF). Here the semantics output, originally expressed in Python dictionary format, is flattened as a text string.}
    \label{fig:model}
\end{figure*}

We tackle the SICSF problem with an end-to-end approach, by using a Conformer-Transformer framework that casts the task as a sequence-to-sequence problem. Based on the intuition that \emph{the SICSF task can be treated as an audio-to-text problem}, we propose to use an encoder pretrained by automatic speech recognition (ASR), which is also an audio-to-text task. The SSL objective, however, focuses on distinguishing one feature from the other features in the same sequence, which is very different from the SICSF task. We propose that, since the ASR objective is closer to the SICSF objective than the SSL objective, ASR-pretrained encoders are more beneficial to the SICSF task than SSL-pretrained encoders. Our main contributions are summarized as follows:
\begin{itemize}
    \item \textbf{Effectiveness}: We present a Conformer-Transformer model with ASR-pretrained encoder that achieves new state-of-the-art performance on the SLURP dataset~\cite{slurp}, outperforming the other end-to-end (E2E) baselines by a large margin. This validates our hypothesis that ASR-pretrained encoders are more suitable for this task than SSL-pretrained encoders because of the task similarity between ASR and SICSF.
    \item \textbf{Efficiency}: We conduct extensive experiments on exploring parameter efficiency of E2E models, including freezing the encoder and using Adapters~\cite{houlsby2019parameter} in the encoder and finetuning on SLURP speech recognition data. Our results show that the best parameter efficiency is only achievable when using ASR-pretrained encoders, while models with SSL-pretrained encoders need finetuning  all parameters to work well.
    \item \textbf{E2E vs. Cascading}: We also compare the proposed end-to-end (E2E) model with cascading models (ASR+NLU), and show that cascading model can only outperform E2E models when using an oracle ASR model. Also, our E2E model with ASR-pretrained encoder is able to match the performance of cascading model with oracle ASR, while all previous E2E models fall behind.
    \item Our code and model checkpoints are open-sourced for use in future research.
\end{itemize}


\section{The Proposed End-To-End Approach}
\subsection{Model}
Our model is based on the encoder-decoder framework. We choose Conformer~\cite{gulati2020conformer} as the encoder since it is one of most popular ASR architectures used in production by many companies (Microsoft~\cite{chen2021continuous}, Google~\cite{gulati2020conformer}). Based on the superior performance of Conformer ASR models, we hypothesize that the Conformer encoder can also be applied to a broader set of audio-to-text tasks such as speech intent classification and slot filling. However, unlike the ASR task that requires monotonic input-output alignment, the speech intent classification and slot filling task is not affected by the orders of predicted entities and thus doesn't require such monotonic property. Therefore, different from the monotonic CTC~\cite{graves2006connectionist} and RNNT~\cite{graves2012sequence} decoders that ASR models usually use, we choose Transformer~\cite{vaswani2017attention} as the decoder (see Figure~\ref{fig:model}), since it has the best global context by letting features of each timestamp to attend to that of all other timestamps.

The intent and slot semantics is formatted as a Python dictionary, which is further flattened as a Python string object. The semantics string is tokenized by SentencePiece~\cite{kudo2018sentencepiece}, while begin-of-sentence (BOS), end-of-sentence (EOS) and padding (PAD) tokens are also added to obtain the target token sequence. In this case, the SICSF task is formulated as an audio-to-text problem similar to speech recognition (ASR). 

\subsection{Training}
We initialize the  encoder with a Conformer-CTC-large model pretrained on NeMo ASR-Set~\footnote{\url{https://catalog.ngc.nvidia.com/orgs/nvidia/teams/nemo/models/stt_en_conformer_ctc_large}}, while the 3-layer Transformer decoder is randomly initialized. 
To circumvent the input-output alignment problem, we follow the seq2seq framework~\cite{sutskever2014sequence} and optimize the model with negative log-likelihood (NLL) loss and teacher-forcing.
Specifically, given the encoder output hidden states as $H=[\mathbf{h}_1,\mathbf{h}_2,...,\mathbf{h}_T]$, and the target sequence labels $Y=[y_1,y_2,...,y_L]$, the training objective is defined as:
\begin{align}
    \mathcal{L} = -\sum_{i=1}^{L-1}\log P(y_{i+1}|y_{i}, y_{i-1},..., y_{1}, H).
\end{align}

\subsection{Inference}

During inference, the input to the decoder is a BOS token and the encoded audio features, and beam search with width 32 and temperature 1.25 is used to obtain the predicted semantics strings, which are then converted to Python dictionaries for evaluation. The strings that have syntax errors during conversion will be treated as empty dictionaries. Missing or invalid values are replaced with ``None'' during conversion.

\section{Experiments}

\subsection{Dataset and Settings}
We use the popular SLURP~\cite{slurp} dataset, which contains around 84 hours of train, 6.9 hours of dev and 10.2 hours of test audios. We use intent classification accuracy and the SLURP metrics proposed in \cite{slurp} for evaluation.

\begin{table*}[t]
\centering
\caption{Comparison with cascading and end-to-end baselines on SLURP~\cite{slurp}. LL60k refers to the LibriLight dataset~\cite{kahn2020librilight}, while LS960 refers to the LibriSpeech dataset~\cite{panayotov2015librispeech}. ``Trainable Params'' refers to the number of parameters that are optimized during training on SLURP dataset. }
\label{tab:results}
\resizebox{0.95\textwidth}{!}{
\begin{tabular}{rrrrrrrr}
\toprule
\multicolumn{1}{c|}{} & \multicolumn{1}{c|}{} & \multicolumn{1}{c|}{} & \multicolumn{1}{c|}{} & \multicolumn{1}{c|}{\textbf{Intent}} & \multicolumn{3}{c}{\textbf{SLURP Metrics}} \\ 
\multicolumn{1}{c|}{\textbf{Model}} & \multicolumn{1}{c|}{\textbf{\begin{tabular}[c]{@{}c@{}} Trainable\\ Params (M)\end{tabular}}} & \multicolumn{1}{c|}{\textbf{\begin{tabular}[c]{@{}c@{}}SSL\\ Pretrained\end{tabular}}} & \multicolumn{1}{c|}{\textbf{\begin{tabular}[c]{@{}c@{}}ASR \\ Pretrained\end{tabular}}} & \multicolumn{1}{c|}{\textbf{Accuracy}} & \multicolumn{1}{c|}{\textbf{Precsion}} & \multicolumn{1}{c|}{\textbf{Recall}} & \multicolumn{1}{c}{\textbf{F1}} \\ 
\hline \hline
\multicolumn{8}{l}{\textbf{Cascading (ASR+NLU)}} \\ \hline
\multicolumn{1}{r|}{Oracle ASR + Transformer} & \multicolumn{1}{r|}{4} & \multicolumn{1}{r|}{None} & \multicolumn{1}{r|}{None} & \multicolumn{1}{r|}{\textbf{90.40}} & \multicolumn{1}{r|}{83.09} & \multicolumn{1}{r|}{80.19} & 81.61 \\ 
\hline
\multicolumn{1}{r|}{Conformer-CTC + Transformer} & \multicolumn{1}{r|}{127} & \multicolumn{1}{r|}{None} & \multicolumn{1}{r|}{\begin{tabular}[c]{@{}r@{}}NeMo ASR-Set,\\ SLURP ASR\end{tabular}} & \multicolumn{1}{r|}{87.18} & \multicolumn{1}{r|}{75.83} & \multicolumn{1}{r|}{72.72} & 74.24 \\ \hline
\multicolumn{1}{r|}{Conformer-CTC + Transformer} & \multicolumn{1}{r|}{4} & \multicolumn{1}{r|}{None} & \multicolumn{1}{r|}{NeMo ASR-Set} & \multicolumn{1}{r|}{71.10} & \multicolumn{1}{r|}{64.49} & \multicolumn{1}{r|}{58.27} & 61.22 \\ \hline \hline
\multicolumn{8}{l}{\textbf{End-to-End}} \\ \hline
\multicolumn{1}{r|}{SpeechBrain-HuBERT-large~\cite{wang2021fine}} & \multicolumn{1}{r|}{317} & \multicolumn{1}{r|}{LL-60k} & \multicolumn{1}{r|}{None} & \multicolumn{1}{r|}{89.37} & \multicolumn{1}{r|}{80.54} & \multicolumn{1}{r|}{77.44} & 78.96 \\ \hline
\multicolumn{1}{r|}{SpeechBrain-HuBERT-base~\cite{wang2021fine}} & \multicolumn{1}{r|}{96} & \multicolumn{1}{r|}{LS-960} & \multicolumn{1}{r|}{None} & \multicolumn{1}{r|}{87.7} & \multicolumn{1}{r|}{77.65} & \multicolumn{1}{r|}{74.78} & 76.19 \\ \hline
\multicolumn{1}{r|}{ESPnet-Conformer~\cite{arora2022espnet}} & \multicolumn{1}{r|}{110} & \multicolumn{1}{r|}{N/A} & \multicolumn{1}{r|}{N/A} & \multicolumn{1}{r|}{86.30} & \multicolumn{1}{r|}{N/A} & \multicolumn{1}{r|}{N/A} & 71.40 \\ \hline
\multicolumn{1}{r|}{Wav2vec-CTI-RoBERTa~\cite{seo2022integration}} & \multicolumn{1}{r|}{200} & \multicolumn{1}{r|}{LS-960} & \multicolumn{1}{r|}{\begin{tabular}[c]{@{}r@{}}LS-960,\\ SLURP ASR\end{tabular}} & \multicolumn{1}{r|}{86.92} & \multicolumn{1}{r|}{N/A} & \multicolumn{1}{r|}{N/A} & 74.66 \\ \hline \hline
\multicolumn{1}{r|}{NeMo-Conformer-Transformer-large} & \multicolumn{1}{r|}{127} & \multicolumn{1}{r|}{None} & \multicolumn{1}{r|}{NeMo ASR-Set} & \multicolumn{1}{r|}{90.14} & \multicolumn{1}{r|}{\textbf{84.31}} & \multicolumn{1}{r|}{\textbf{80.33}} & \textbf{82.27} \\ 
\bottomrule
\end{tabular}}
\end{table*}

\subsection{Implementation Details}
Our model is implemented with PyTorch and NeMo\footnote{\url{https://github.com/NVIDIA/NeMo}}. We set the tokenizer vocabulary size to 58, and each token is embedded as a 512-dimension feature. We use Adam optimizer, with momentum $[0.9,0.98]$ and no weight decay. The learning rates for encoder and decoder are 2e-4 and 3e-4 respectively. A Cosine annealing scheduler with 2000 linear warm-up steps is applied. The proposed model and its variants are trained on an NVIDIA RTX A6000 GPU with batch size 32 for 100 epochs. 

For the NLU model in cascading baselines, we use a vocabulary size of 1024 for input text from ASR, while the output vocabulary size is 512. The ASR model in cascading baseline is initialized as the same in our E2E SLU model. Before finetuning on SLURP ASR, the ASR model has word-error-rate (WER) of 23.5 on SLURP test set. The WER decreased to 9.5 after finetuning on SLURP ASR data. 


\subsection{Comparison with Baselines}
We compare the proposed model with three E2E baselines: SpeechBrain-HuBERT~ \cite{wang2021fine}, ESPnet-Conformer ~\cite{arora2022espnet} and Wav2Vec-CTI-RoBERTa~\cite{seo2022integration}. We also compare with two cascading (ASR+NLU) baselines, where the ASR model is a Conformer-CTC-large mode finetuned on SLURP, and the NLU model is a Transformer with 3 encoder and 3 decoder layers. Another cascading baseline with oracle ASR is also included.

We show the main results in Table~\ref{tab:results}. As we can see, the cascading baseline with oracle ASR works better than the other SLU baselines, as is also observed in previous works~\cite{slurp,seo2022integration}. However, our E2E SLU model, with 82.27\% SLURP-F1, is able to match the performance of cascading models with oracle ASR, and outperforms the other E2E SLU models by a noticeable margin. It can also be noted that our model with 127M parameters is able to achieve 2.3\% higher SLURP-F1 than the second best SLU model with 317M parameters. Compared with Wav2vec-CTI-RoBERTa~\cite{seo2022integration}, our model does not require finetuning on SLURP ASR, which makes our model more efficient to train. Overall, the superior performance of our model suggests that using an encoder pretrained on large ASR dataset is much more beneficial than encoders pretrained by self-supervised learning in speech intent classification and slot filling (SICSF). We will discuss this finding with more details in Section~\ref{sec:asr-ssl}.

The cascading baseline, when not finetuned with ASR on SLURP, has much lower performance than the other baselines. Meanwhile, when the cascading baseline is finetuned on SLURP ASR, it's able to reach better performance than ESPnet-Conformer~\cite{arora2022espnet} and comparable slot filling performance with Wav2vec-CTI-RoBERTa~\cite{seo2022integration}. This shows that, cascading ASR+NLU models, although have the drawback of propagating errors from ASR to NLU, still have the potential of getting good performance.


\begin{table}[t]
\centering
\caption{Study on parameter efficiency. All models use the same Conformer-Transformer architecture.}
\label{tab:effect}
\resizebox{\columnwidth}{!}{
\begin{tabular}{|r|r|r|r|r|}
\toprule
\multicolumn{1}{|c|}{\textbf{\begin{tabular}[c]{@{}c@{}}Freeze\\ Encoder\end{tabular}}} & \multicolumn{1}{c|}{\textbf{\begin{tabular}[c]{@{}c@{}}Use\\ Adapter\end{tabular}}} & \multicolumn{1}{c|}{\textbf{\begin{tabular}[c]{@{}c@{}}Trainable\\ Params (M)\end{tabular}}} & \multicolumn{1}{c|}{\textbf{\begin{tabular}[c]{@{}c@{}}Pretrained\\ (Task: Dataset)\end{tabular}}} & \multicolumn{1}{c|}{\textbf{SLURP-F1}} \\ \hline \hline
No & No & 127 & SSL: LL60kh & 77.22 \\ \hline
Yes & No & 12 & SSL: LL60kh & 36.21 \\ \hline
Yes & Yes & 13 & SSL: LL60kh & 43.28 \\ \hline \hline
No & No & 127 & ASR: NeMo ASR-Set & \textbf{82.27} \\ \hline
Yes & No & 12 & ASR: NeMo ASR-Set & 72.59 \\ \hline
Yes & Yes & 13 & ASR: NeMo ASR-Set & 77.54 \\ 
\bottomrule
\end{tabular}}
\end{table}

\begin{table}[t]
\centering
\caption{Ablation study on effectiveness of pretraining. All models use the same Conformer-Transformer-large network.}
\label{tab:pretrain}
\resizebox{0.9\columnwidth}{!}{
\begin{tabular}{|r|r|rrr|}
\toprule
\multicolumn{1}{|c|}{\textbf{}} & \multicolumn{1}{c|}{\textbf{Intent}} & \multicolumn{3}{c|}{\textbf{SLURP Metrics}} \\ 
\multicolumn{1}{|c|}{\textbf{\begin{tabular}[c]{@{}c@{}}Pretrained\\ (Task: Dataset)\end{tabular}}} & \multicolumn{1}{c|}{\textbf{Accuracy}} & \multicolumn{1}{c|}{\textbf{Precision}} & \multicolumn{1}{c|}{\textbf{Recall}} & \multicolumn{1}{c|}{\textbf{F1}} \\ 
\hline 
\hline
ASR: NeMo ASR-Set & \textbf{90.14} & \multicolumn{1}{r|}{\textbf{84.31}} & \multicolumn{1}{r|}{\textbf{80.33}} & \textbf{82.27} \\ \hline
ASR: LS960h & 92.17 & \multicolumn{1}{r|}{81.15} & \multicolumn{1}{r|}{77.33} & 79.19 \\ \hline
SSL: LL60kh & 89.40 & \multicolumn{1}{r|}{77.90} & \multicolumn{1}{r|}{76.65} & 77.22 \\ \hline
None & 72.56 & \multicolumn{1}{r|}{53.59} & \multicolumn{1}{r|}{53.92} & 53.76 \\ \bottomrule
\end{tabular}}
\end{table}

\begin{figure}[t]
    \centering   \includegraphics[width=0.8\linewidth]{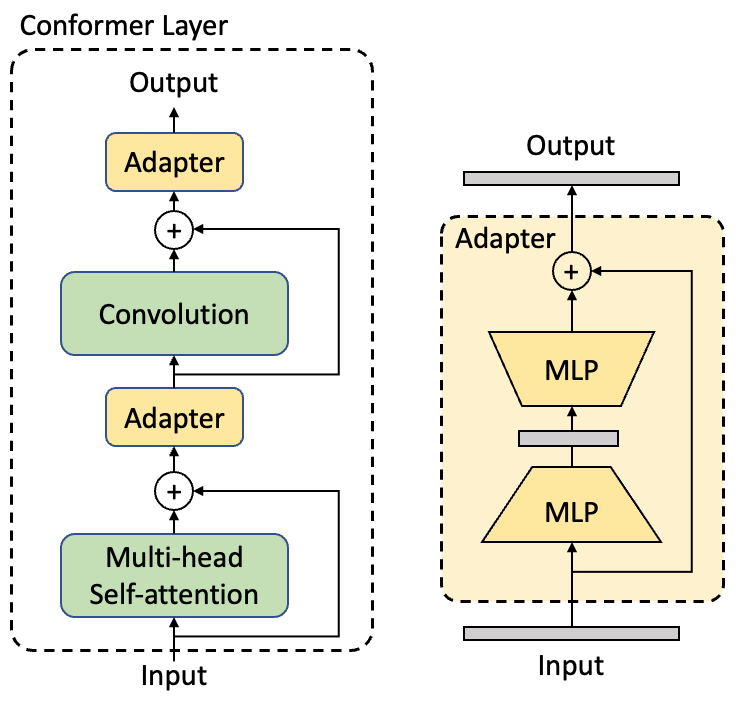}
    \caption{\textbf{Left}: Illustration of adding Adapter~\cite{houlsby2019parameter} modules to a Conformer~\cite{gulati2020conformer} layer, where multi-layer perceptrons (MLPs) and layer-normalization are omitted for clarity. \textbf{Right}: Illustration of an Adapter module, where non-linear activation is omitted for clarity.}
    \label{fig:adapter}
\end{figure}

\subsection{Exploring Parameter Efficiency}
Adapters~\cite{houlsby2019parameter} were proposed to improve models' parameter efficiency in transfer learning for NLP, by adding only about 1\% of the whole network parameters to the frozen model while achieving performance comparable to full finetuning. Each Adapter consists of several multi-layer perceptrons with very small intermediate dimensions (e.g., 32) and residual connections (Figure~\ref{fig:adapter}-Right). Since Adapters are proven to be very effective in speech recognition as well~\cite{thomas2022efficient,sathyendra2022contextual}, here we also study whether Adapters can help the SICSF task. As shown in Figure~\ref{fig:adapter}-Left, Adapters are added to each Conformer layer in the encoder of our model, while the encoder is frozen during training. We apply Adapters to both ASR-pretrained and SSL-pretrained encoders, and show the results in Table~\ref{tab:effect}. 

As we can see, for the SSL-pretrained encoder, merely freezing it without adding Adapters leads to very low performance, which is 40\% absolute decrease from training the full model. By adding about 1M parameters, Adapters are able to improve the performance of SSl-pretrained encoders from 36.21\% to 43.28\%, which is about 7\% improvement. On the other hand, merely freezing the ASR-pretrained encoder has 72.59\% SLURP-F1, which is almost double the performance of SSL-pretrained encoder. This shows that the ASR-pretrained encoder itself is already able to generate representative features for the SICSF task, even without any finetuning. Meanwhile, adding Adapters to the frozen ASR-pretrained encoder only improves the performance by around 5\%, which is an even smaller improvement than it is in the case of SSL-pretrained encoder. However, we can also see that adding Adapters to frozen ASR-pretrained encoder is able to match the performance of training the full model with SSL-pretrained encoder, and also outperforms some baselines~\cite{arora2022espnet, seo2022integration} in Table~\ref{tab:results}. Although using Adapters in frozen ASR-pretrained encoder still cannot match the performance of finetuning the whole encoder, considering the size of the model, it still achieves a good balance between performance and network size. From these results, we can conclude that having better parameter efficiency while maintaining good performance is only achievable for ASR-pretrained encoders, while model with SSL-pretrained encoder still needs a lot of parameters to work well, since the SICSF task is closer to audio-to-text in ASR rather than the frame discrimination task in SSL.

\subsection{Ablation Study}

\subsubsection{Effect of Pretraining: ASR vs. SSL}
\label{sec:asr-ssl}

We also try using SSL-pretrained encoder~\footnote{\url{https://catalog.ngc.nvidia.com/orgs/nvidia/teams/nemo/models/ssl_en_conformer_large}}, and compare the results with ASR-pretrained encoders. We also add a baseline without any pretraining, and show the results in Table~\ref{tab:pretrain}. As we can see, training from scratch has almost 30\% decrease from the best model, while SSL-pretrained encoder with 77.22\% SLURP-F1 lies in between. This shows that pretraining is still essential to the model's performance. We also try using ASR encoder pretrained on the LibriSpeech~\cite{panayotov2015librispeech} dataset, which has about 2\% higher SLURP-F1 than SSL-pretrained encoder. Considering the fact that LibriLight~\cite{kahn2020librilight} has 60k hours audio, while LibriSpeech~\cite{panayotov2015librispeech} only has 960 hours, we can see that ASR-pretrained encoder is more efficient in utilizing pretraining data than SSL-pretrained encoder. In addition, we try a Conformer-Xlarge encoder with 5x the size of Conformer-large, pretrained on LL60k~\cite{kahn2020librilight}, and find that it's only with this extra large model that using SSL-pretrained encoder can have a similar performance (SLURP-F1=81.28\%) as using ASR-pretrained encoder (SLURP-F1=82.27\%). In other words, using ASR-pretrained encoders is also more cost-efficient in terms of model size.

\subsubsection{Effect of Finetuning on SLURP ASR}
We further explore the effect of finetuning the ASR-pretrained encoder on SLURP ASR, and found that the performance does not change much. The intent accuracy and SLURP-F1 for the finetuned model are 90.28\% and 82.08\%, while with the ones without finetuning on SLURP ASR are 90.14\% and 82.27\%. The reason for the similar performance is that the encoder pretrained on large ASR datasets already has the knowledge of audio-to-text task, while finetuning on SLURP ASR only improves the model's knowledge on dataset statistics. However, since later training on the SICSF task can also help the model learn dataset statistics, the effect of finetuning on SLURP ASR is less obvious.

\begin{table}[t]
\centering
\caption{Ablation study on output vocabulary size. All models use the same Conformer-Transformer architecture.}
\label{tab:vocab}
\resizebox{0.8\columnwidth}{!}{
\begin{tabular}{|r|r|rrr|}
\toprule
\multicolumn{1}{|c|}{\textbf{}} & \multicolumn{1}{c|}{\textbf{Intent}} & \multicolumn{3}{c|}{\textbf{SLURP Metrics}} \\ \hline
\multicolumn{1}{|c|}{\textbf{Vocab}} & \multicolumn{1}{c|}{\textbf{Accuracy}} & \multicolumn{1}{c|}{\textbf{Precision}} & \multicolumn{1}{c|}{\textbf{Recall}} & \multicolumn{1}{c|}{\textbf{F1}} \\ \hline
58 & 90.14 & \multicolumn{1}{r|}{\textbf{84.31}} & \multicolumn{1}{r|}{\textbf{80.33}} & \textbf{82.27} \\ \hline
256 & 90.12 & \multicolumn{1}{r|}{83.09} & \multicolumn{1}{r|}{79.47} & 81.24 \\ \hline
512 & 90.13 & \multicolumn{1}{r|}{83.19} & \multicolumn{1}{r|}{79.77} & 81.44 \\ \hline
1024 & \textbf{90.17} & \multicolumn{1}{r|}{82.31} & \multicolumn{1}{r|}{79.45} & 80.86 \\ 
\bottomrule
\end{tabular}}
\end{table}

\subsubsection{Effect of Vocabulary Size}
As performance of Confomer-based ASR models are usually affected by vocabulary sizes, where CTC~\cite{graves2006connectionist} decoder usually works better with smaller (e.g., 128) vocabulary while RNNT~\cite{graves2012sequence} decoder is better with larger ones (e.g., 1024), here We study the effect of different vocabulary size on the proposed model, and show the results in Table~\ref{tab:vocab}. The intent accuracy remains pretty stable with different vocabulary sizes, while the best SLURP-F1 is obtained with the smallest vocabulary size. This is different from what we have observed for cascading models, where their performance grows as the output vocabulary size grows, and saturates around size 512. It should also be noted that cascading models perform badly with small input vocabulary size (e.g., 58), which is because the input has more diverse natural language while the output semantics has a more limited set of words.

\section{Conclusion}
We present a Conformer-Transformer model for end-to-end speech intent classification and slot filling (SICSF), where the Conformer encoder is pretrained on a large collection of speech recognition (ASR) datasets. Our model is able to achieve new stat-of-the-art results on the SLURP dataset. We also compare with cascading models, and show that our model can match the performance of cascading model with oracle ASR, while previous end-to-end models fall behind. We also study the effect of encoders pretrained by self-supervised learning (SSL), and show that  ASR-pretrained encoder achieves noticeably better performance, since the SICSF objective is more similar to ASR than SSL. We also explore the parameter efficiency of our model, and show that using adapters in frozen ASR-pretrained encoder can still achieve very good performance, while SSL-pretrained encoder needs full finetuning to work well. Our code and checkpoints are publicly available to benefit future research.

\bibliographystyle{IEEEtran}
\bibliography{refs}

\end{document}